\documentclass[10pt,letterpaper]{article}

\usepackage{cogsci}

\cogscifinalcopy 

\usepackage[
    backend=biber,
    style=apa,
    natbib=true,
    doi=false,
    isbn=false,
    url=false,
]{biblatex}
\usepackage{pslatex}
\usepackage{float} 
\usepackage{tabularx}%
\usepackage{stfloats}%
\usepackage{float}
\usepackage{booktabs} 
\usepackage{multirow}%
\usepackage{makecell} 
\usepackage{graphicx}
\usepackage{hyperref}
\usepackage{amsmath}
\usepackage{balance}
\usepackage{fontawesome5}

\title{Do Large Language Models Truly Grasp Mathematics? \\An Empirical Exploration from Cognitive Psychology}
 
\author{ \bf Shuoyoucheng Ma\textsuperscript{1,†},Wei Xie\textsuperscript{1,†,\faIcon{envelope}},  Zhenhua Wang\textsuperscript{1},Xiaobing Sun\textsuperscript{2}, Kai Chen\textsuperscript{3}\\
\bf Enze Wang\textsuperscript{1},Wei Liu\textsuperscript{1},Hanying Tong\textsuperscript{1} \\
\\
\textsuperscript{1}College of Computer Science and Technology, National University of Defense Technology\\
\textsuperscript{2}Institute of High Performance Computing, Agency for Science, Technology and Research (A*STAR)\\
\textsuperscript{3}Institute of Information Engineering, Chinese Academy of Sciences\\
\textsuperscript{\faIcon{envelope}}Corresponding author: xiewei@nudt.edu.cn\\
\textsuperscript{†}Equal contributors
}

\begin{document}

\maketitle

\begin{abstract}
The cognitive mechanism by which Large Language Models (LLMs) solve mathematical problems remains a widely debated and unresolved issue. Currently, there is little interpretable experimental evidence that connects LLMs' problem-solving with human cognitive psychology.
To determine whether LLMs possess human-like mathematical reasoning, we modified the problems used in the human Cognitive Reflection Test (CRT). Our results show that even with the use of Chain-of-Thought (CoT) prompts, mainstream LLMs, including the o1 model (noted for its reasoning capabilities), have a high error rate when solving these modified CRT problems. Specifically, the average accuracy rate dropped by up to 50\% compared to the original problems.
Further analysis of LLMs' incorrect answers suggests that they primarily rely on pattern matching from their training data, which aligns more with human intuition (System 1 thinking) rather than with human-like reasoning (System 2 thinking). This finding challenges the belief that LLMs have genuine mathematical reasoning abilities comparable to humans. As a result, this work may adjust overly optimistic views on LLMs' progress toward Artificial General Intelligence. Our dataset and experimental data can be accessed at \href{https://osf.io/74yj2/}{\url{https://osf.io/74yj2/}}.

\textbf{Keywords:} 
Larger Language Model, Chain-of-Thought, Cognitive Reflection Test
\end{abstract}

\section{Introduction}


Large Language Models (LLMs) are considered the dawn of Artificial General Intelligence (AGI) (\cite{bubeck_sparks_2023}). Models such as ChatGPT (\cite{openal_introducing_nodate}), GPT-4 (\cite{bubeck_sparks_2023}), Claude (\cite{anthropic_introducing_nodate}), Gemini (\cite{gemini_team_gemini_2024}), GLM (\cite{du_glm_nodate}), and o1-preview (\cite{o1_system_card}) have garnered considerable attention from both academia and industry. These models exhibit significant potential across various fields, including education (\cite{perez-nunez_exploring_2023}), healthcare (\cite{egger_chatgpt_2024,ward_healai_2024}), coding (\cite{yu_wavecoder_2024}), and social governance (\cite{azim_framework_2024}). This is partly attributed to the `emergence phenomenon' (\cite{schaeffer_are_2023}), which allows LLMs, due to their large training datasets and numerous parameters, to perform tasks they were not specifically trained for. 

Using mathematical skills as an example, most LLMs have demonstrated remarkable abilities to tackle these problems. Using the Chain-of-Thought (CoT) method, the capacity of LLMs to solve mathematical problems can be further increased (\cite{wei_chain_nodate,bi_stoc-tot_nodate,wang_self-consistency_nodate,yao_tree_nodat,brown_language_2020}). However, due to the interpretability challenges posed by large-scale neural networks, there is still no scientific consensus on the origins and mechanisms underlying the mathematical capabilities of LLMs.

Conducting psychological measurement experiments on LLMs can help enhance the interpretability of research into LLMs' thinking. Previous research (\cite{hagendorff_human-like_2023}) has demonstrated through experiments that the CoT method can effectively help LLMs handle the pitfalls in Cognitive Reflection Test (CRT) problems (\cite{frederick_cognitive_2005}). CRT problems are some well-crafted math or logic problems that human testers often get wrong(\cite{hagendorff_human-like_2023}) due to intuitive thinking (System 1(\cite{kahneman_judgment_1974,sloman_empirical_1996,kahneman_thinking_2015,stanovich_who_1999}) ). By using the CoT method, LLMs are prompted to rely more on human-like logical reasoning (System 2(\cite{kahneman_judgment_1974,sloman_empirical_1996,kahneman_thinking_2015,stanovich_who_1999})), enhancing their accuracy in solving these problems. Advanced models like GPT-4 have even achieved higher accuracy than humans in these CRT tasks. However, the authors also raised speculative questions in their work, suggesting, ``It is possible that some models encountered enough examples in their training to solve them `from memory'".

We repeated and improved the experiment conducted by the previous study. If LLMs genuinely possess the intrinsic capability to comprehend mathematical logic, as is widely hypothesized, their accuracy in responding to the modified problems should not experience a marked decrement. However, our experiments revealed a distinctly opposing result: Even when employing the CoT approach, prominent LLMs, including the latest o1 model, continued to manifest a considerable error rate for the modified problems. Further analysis of the incorrect answers indicates that LLMs may not have developed a logical reasoning proficiency akin to System 2 or acquired comprehensive mathematical cognitive abilities. They predominantly resort to a methodology reminiscent of System 1, which involves matching and producing responses based on the similarity between user inquiries and training data during the text generation process. This investigation may serve to temper the overly optimistic anticipations regarding the effectiveness of CoT and the competencies of LLMs in approximating AGI.

\section{Experiment Design}\label{sec2}

\textbf{Method for Problem Modification.} A single researcher manually performed all modifications to the dataset in our study. To ensure the accuracy of these modifications, we also invited an additional researcher to verify the problems.


\textbf{Method for Experimental Implementation.} To uncover the problem-solving strategies of LLMs, we appended explicit instructions (`Please think step by step.') to the end of each problem, encouraging the use of the CoT method.


\textbf{Methods for Result Analysis and Statistics.} 
In this study, we comprehensively evaluated five prominent open-source and commercial LLMs. To ensure the accuracy of the assessment results, two independent researchers performed a detailed analysis and documentation of each model’s responses to all queries. A response was considered accurate only when the problem-solving approach and the final answer were correct. After the statistical examination by the two independent researchers, a third researcher reviewed the discrepancies in their analytical results and made a final determination to ensure consistency in the results. Furthermore, to enhance the reliability of the experimental findings, three replicate trials were conducted on each dataset, with average values subsequently calculated. This methodological framework carried out all experimental procedures described in this study.


\textbf{Method for Human Comparative Experimentation.}
We selected five researchers not involved in this study in our laboratory for this Experimentation. Each individual responded to 20\% of the problems from each dataset, constituting a comprehensive examination of all datasets. We required them to provide the problem-solving process (activating system 2). The scoring methodology was consistent with that employed for LLMs. We calculated the accuracy based on these five examination papers, which served as a control group for comparison with the LLMs' results. 


\section{Experiment I: Changing the Numbers in A Problem Without Altering Its Description and Principle}\label{sec2}
We first replicated the experiment on the CRT3 data set from the study by Hagendorff(\cite{hagendorff_human-like_2023}), which comprises 50 mathematical problems. We introduced three types of modifications to test problems, as delineated in Table \ref{tab:my_tab1}. Type A represents the original problems in the CRT3 test, characterized as typical exponential growth problems, and incorporates three numbers in the problem statement. Type B modifies two of these numbers, while type C modifies all three. Type D replaces two numbers with letters, transforming the arithmetic problem into an algebraic one. To prevent LLMs from being unsure how to handle algebraic symbols, we appended a suffix for guidance: ``X and Y are both numbers, you can use them to represent the final answer." Modifications of Types B, C, and D do not alter the description of the original problem (Type A). Thus, the underlying mathematical principle of the original problem remains unchanged.

\begin{figure}[H]
\centering
\includegraphics[width=0.49\textwidth]{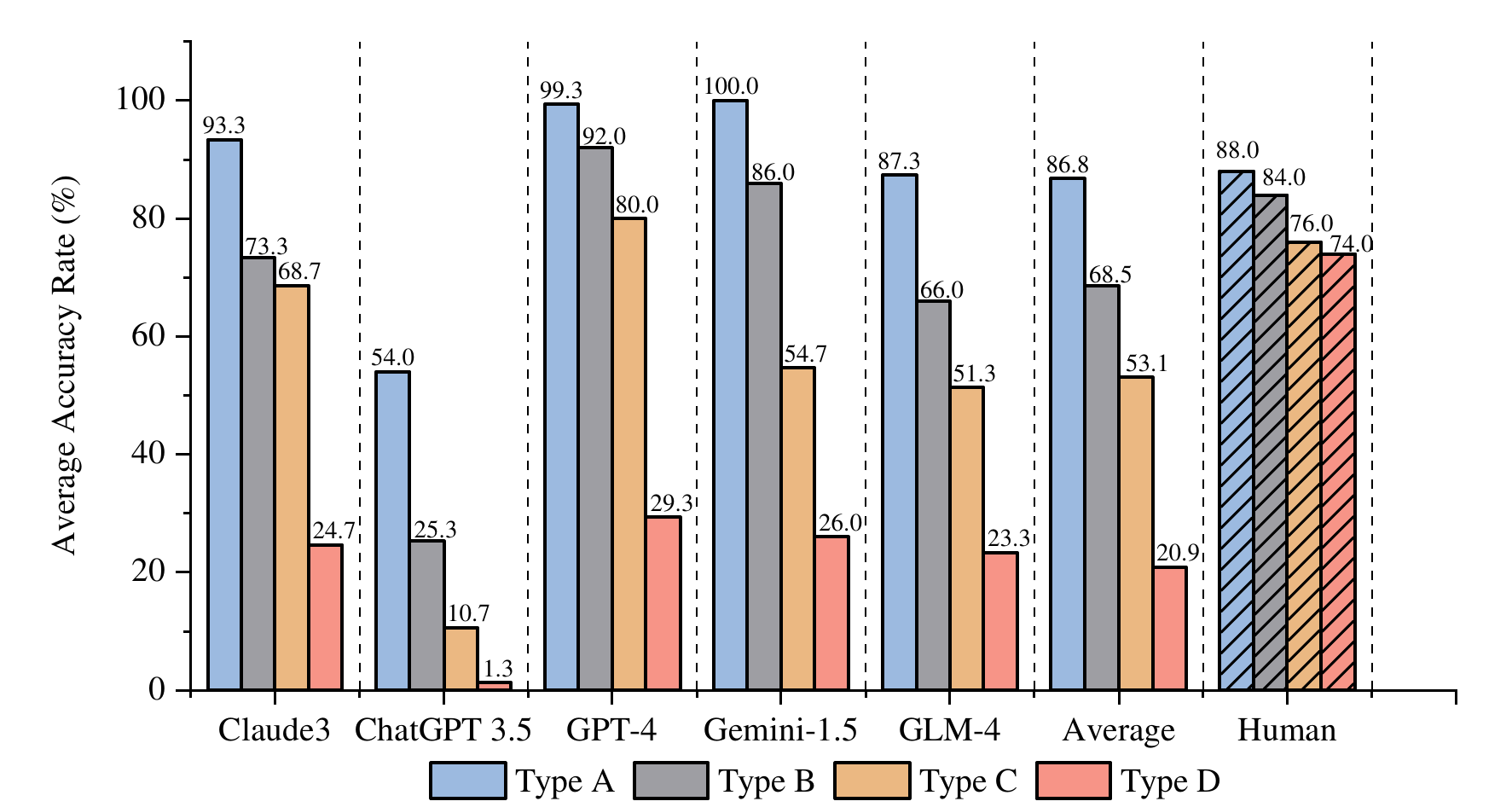}

\caption{Accuracy of five mainstream LLMs in answering four types of problems. Human experimental results are indicated with slashes.}\label{fig:myfig1}
\end{figure}

The results are shown in Fig.\ref{fig:myfig1}. In the case of Type A problems, the highest accuracy reached 100\%, the lowest was 54.0\%, and the average accuracy was 86.8\%. For Type B problems, which involved modifications to two numbers, the accuracy significantly decreased, with the highest being 92.0\%, the lowest 25.3\%, and the average 68.5\%($\delta = 18.3\%; \chi^2(1) = 23.00; P < 0.001$). For the problems of Type C, where three numbers were modified, the accuracy further declined, with the highest at 80.0\%, the lowest at 10.7\%, and the average 53.1\%($\delta = 15.4\%; \chi^2(1) = 11.91; P < 0.001$). The accuracy again dropped sharply when the arithmetic problems were transformed into algebraic ones (Type D). The best-performing model, GPT-4, achieved an accuracy of only 29.3\%, whereas ChatGPT 3.5 scarcely provided correct answers to any problems. The average accuracy was reduced to 20.9\%($\delta = 32.2\%; \chi^2(1) = 54.00; P < 0.001$). However, the accuracy of human participants did not show a significant decrease (Fig. \ref{fig:myfig1}).


\begin{table}[htp]
\centering
\caption{Problem Types and Modifications}
\vskip 0.12in
\label{tab:my_tab1}
\begin{tabularx}{\linewidth}{p{0.5cm}p{1.9cm}X}
\toprule
\textbf{Type} & \textbf{Modification} & \textbf{Example} \\
\midrule
\multirow{8}{*}{A} & \multirow{8}{*}{\makecell{  Original \\ Problem}} & 
In a city, a virus is spreading, causing the total number of infected individuals to \textbf{double each day}. If it takes \textbf{6 days} for the entire city's population to be infected, how many days would it require for \textbf{half of} the people to become infected? Please think step by step. \\
\midrule
\multirow{8}{*}{B} & \multirow{8}{*}{\makecell{Two \\Numbers\\ Modified}} & 
In a city, a virus is spreading, causing the total number of infected individuals to \textbf{triple each day}. If it takes 6 days for the entire city's population to be infected, how many days would it require for \textbf{1/27 of} the people to become infected? Please think step by step. \\
\midrule
\multirow{8}{*}{C} & \multirow{8}{*}{\makecell{All Three\\ Numbers\\ Modified}} & 
In a city, a virus is spreading, causing the total number of infected individuals to \textbf{triple every three days}. If it takes \textbf{36 days} for the entire city's population to be infected, how many days would it require for \textbf{1/27} of the people to become infected? Please think step by step. \\
\midrule
\multirow{10}{*}{D} & \multirow{10}{*}{\makecell{Algebraic\\ Trans- \\formation}} & 
In a city, a virus is spreading, causing the total number of infected individuals to \textbf{triple every 3X days}. If it takes \textbf{6Y days} for the entire city's population to be infected, how many days would it require for 1/27 of the people to become infected? X and Y are both numbers, and you can use them to represent the final answer. Please think step by step. \\
\bottomrule
\end{tabularx}
\end{table}

\begin{figure}[htb]
\centering
\includegraphics[width=0.5\textwidth]{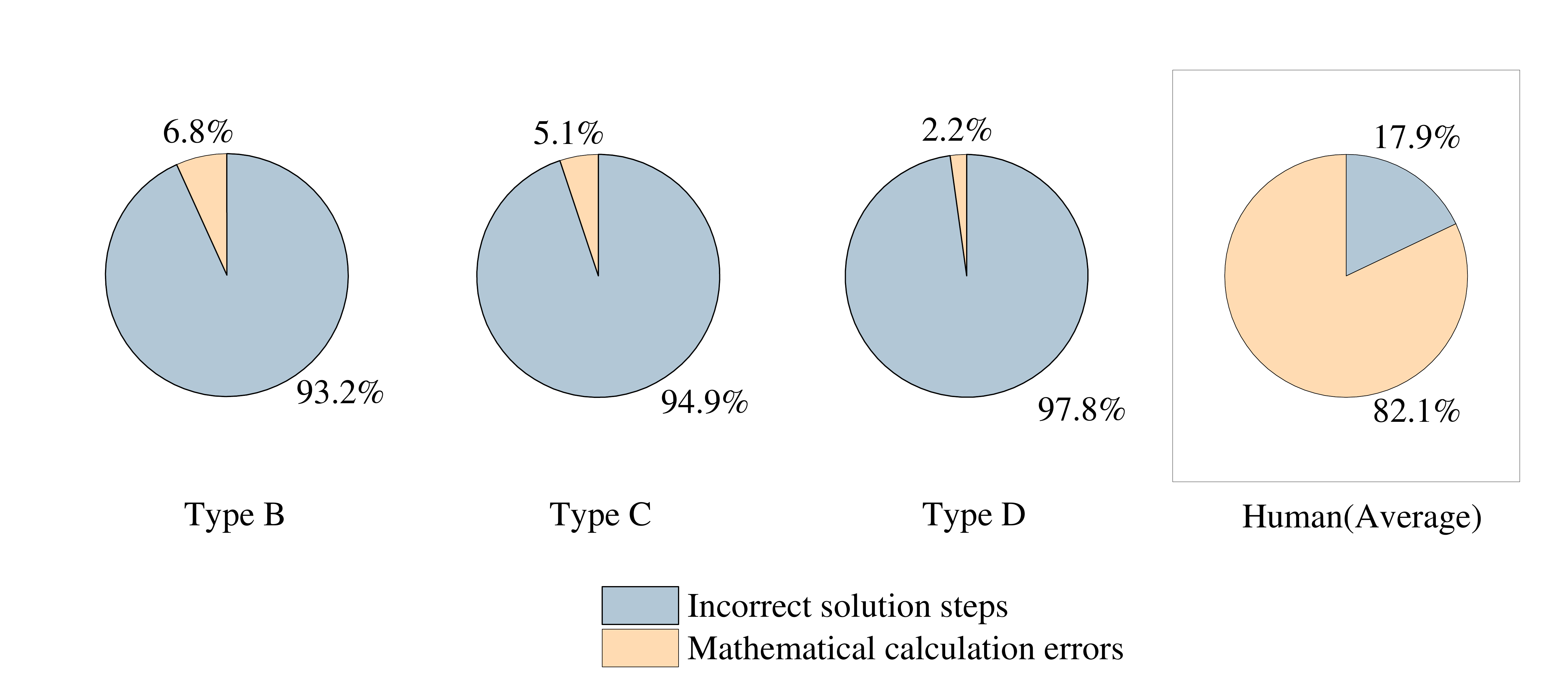}

\caption{The proportion of incorrect answer types in answering three types of problems. Human experimental results are indicated with box.}\label{Fig:fig_result_1-2}
\end{figure}

As shown in Fig. \ref{Fig:fig_result_1-2}, we analyzed all incorrect answers provided by the five LLMs for each type of problem and found that in type B, errors arising from incorrect solution steps (e.g., omitting a step or altering the original calculation method) constituted 93.2\%. In contrast, those only due to mathematical calculation (e.g., 16/2=4) accounted for 6.8\%. In type C, errors due to incorrect solution steps accounted for 94.9\%, and errors solely due to mathematical calculation accounted for 5.1\%. In type D, errors due to incorrect solution steps accounted for 97.8\%, and errors caused solely by mathematical calculation accounted for 2.2\%. In contrast, the majority (82.1\%) of human errors are found in mathematical calculation (Fig. \ref{Fig:fig_result_1-2}). Post-analysis discussions with test-takers revealed that these errors primarily stem from the extensive number of problems, causing calculative exhaustion.


\textbf{Result Analysis of Experiment I:} The modifications in types B, C, and D only altered the specific numbers in the original problems without changing the problems' mathematical principles and computational rules. For a human with mathematical and logical thinking who can solve problems by formulating equations or programming, merely changing the input numbers without modifying the mathematical principles should not significantly decrease accuracy (as shown in Fig. \ref{fig:myfig1}). However, the performance of LLMs differs significantly from that.

\section{Experiment II: Modifying the Principle of A Problem While Maintaining Similarity in Its Description}\label{sec3}

We designed a reverse experiment to further substantiate the inferences drawn from Experiment I. In this experiment, we substantially altered the fundamental principles of the mathematical problems, endeavoring to preserve descriptions that closely mirrored the original versions. Subsequently, we investigated whether LLMs persisted in utilizing their problem-solving strategies on the original problem or adapted their methods to the modified problem throughout the problem-solving process. We conducted three distinct subexperiments corresponding to the three types of datasets.

\subsection{For CRT1: Transforming Additively Separable Problems into Non-Additively Separable Problems }\label{subsec3_1}
\begin{figure}[H]
\centering
\includegraphics[width=0.5\textwidth]{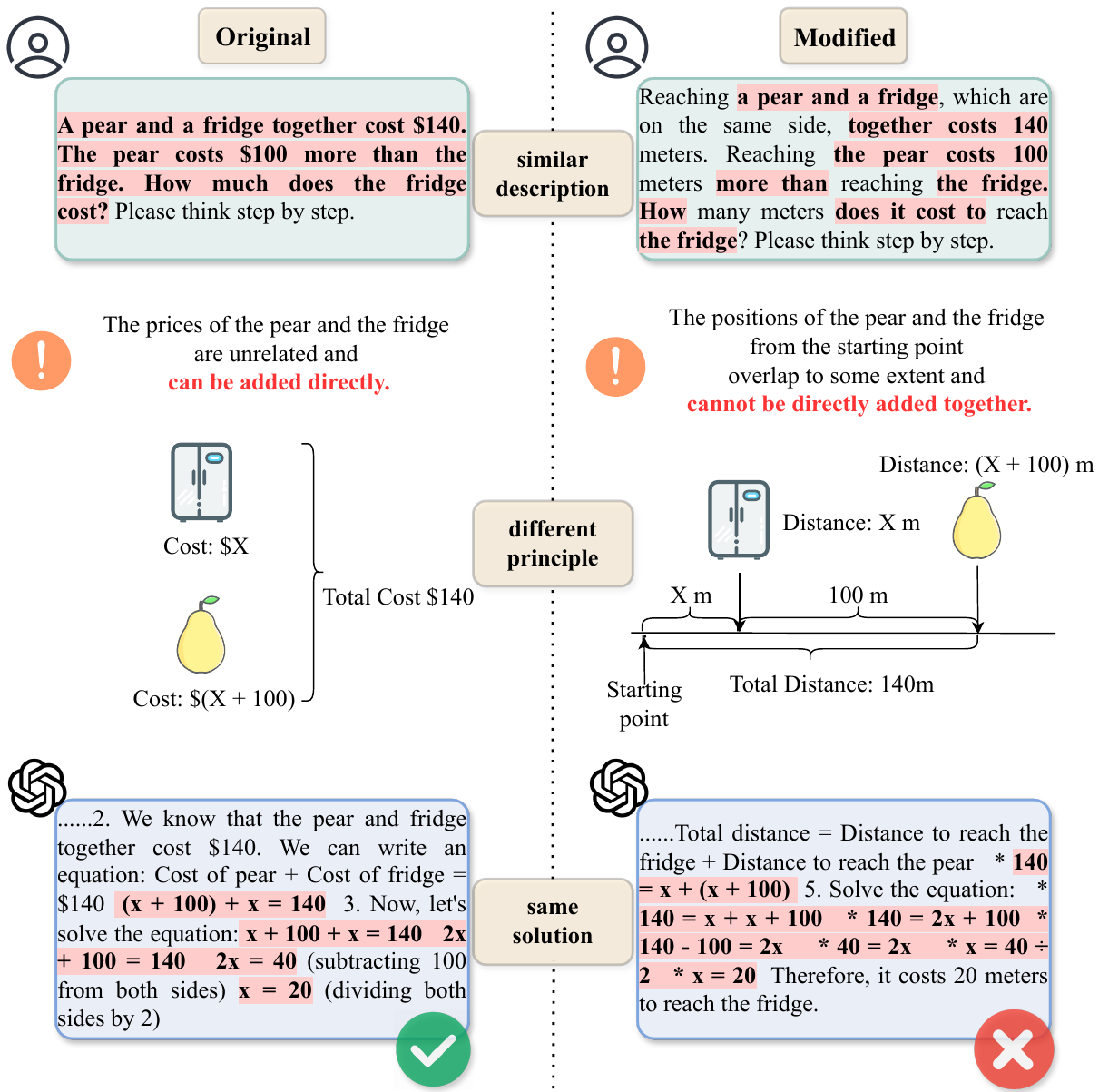}
\caption{An example of an original CRT1 problem and its modified problem.}\label{Fig:fig2-1}
\end{figure}

For instance, Fig. \ref{Fig:fig2-1} illustrates the crucial distinction between an original CRT1 problem and its modified version. In the original problem, the total cost is simply the sum of the prices of two items. However, in the modified version, reaching the two items from the starting point through a shared segment must be deducted from the overall distance calculation. We use the string similarity calculation function within the difflib module of Python to quantify the degree of similarity in textual expression between the modified problem and the original, resulting in an average similarity of 75.91\%. Despite the similar wording in the problem statements, the underlying mathematical principles are fundamentally distinct.

\begin{figure}[H]
\centering
\includegraphics[width=0.48\textwidth]{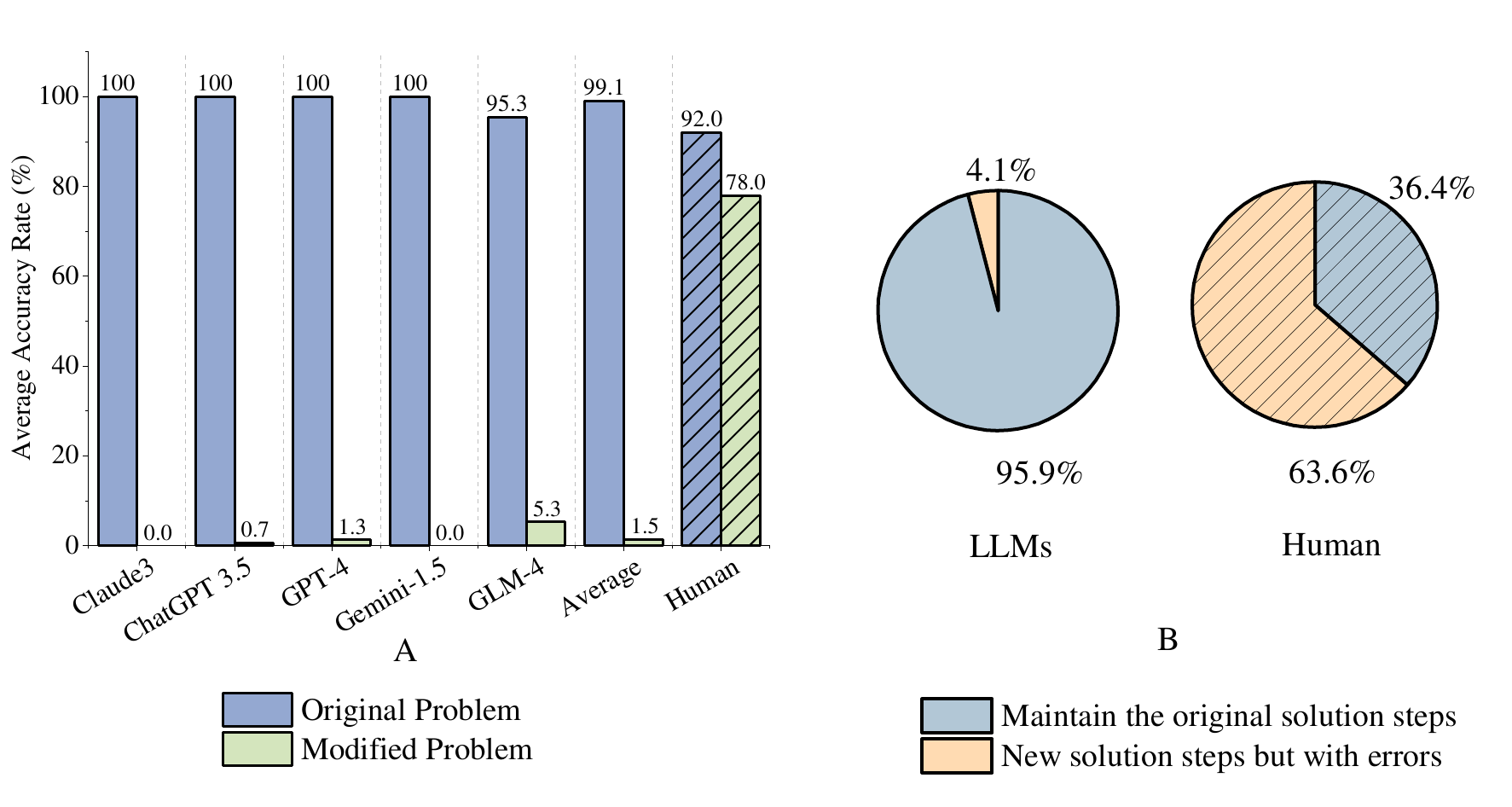}
\caption{Accuracy of the LLMs and Human(with slash) when answering the CRT1 problems (A), and the proportion of incorrect answer types in modified CRT1 problems (B).} \label{Fig:fig_result_2-1}
\end{figure}

Fig. \ref{Fig:fig_result_2-1}A illustrates our findings. For the original CRT1 problems, four LLMs achieved a perfect accuracy of 100\%, with GLM-4 closely following at 95.3\%, yielding an overall accuracy of 99.1\%. However, when confronted with the modified problems, the accuracy significantly declined. GLM-4 recorded the highest accuracy rate, albeit a mere 5.3\%. Claude3 and Gemini-1.5 failed to provide correct answers to any of the modified problems. On average, the five LLMs attained only 1.5\% accuracy($\delta = 97.6\%; \chi^2(1) = 472.41; P < 0.001$). Human accuracy has decreased by only 14\%. This means that problems modified to include overlapping components have become more complex and require more thought.

By analyzing incorrect answers, as shown in Fig. \ref{Fig:fig_result_2-1}B, we found that 95.9\% of errors were attributed to the application of the original problem's solution method to the modified problem (in the problem-solving process, add the distance directly.), whereas 4.1\% resulted from new solution steps but with errors. Relatively speaking, human experimenters less frequently adopt the original problem's solution method.

\subsection{For CRT2: Transforming Problems Related to the Number of Individuals into Problems Independent of Individual Count}\label{subsec3_2}

\begin{figure}[htp]
\centering
\includegraphics[width=0.48\textwidth]{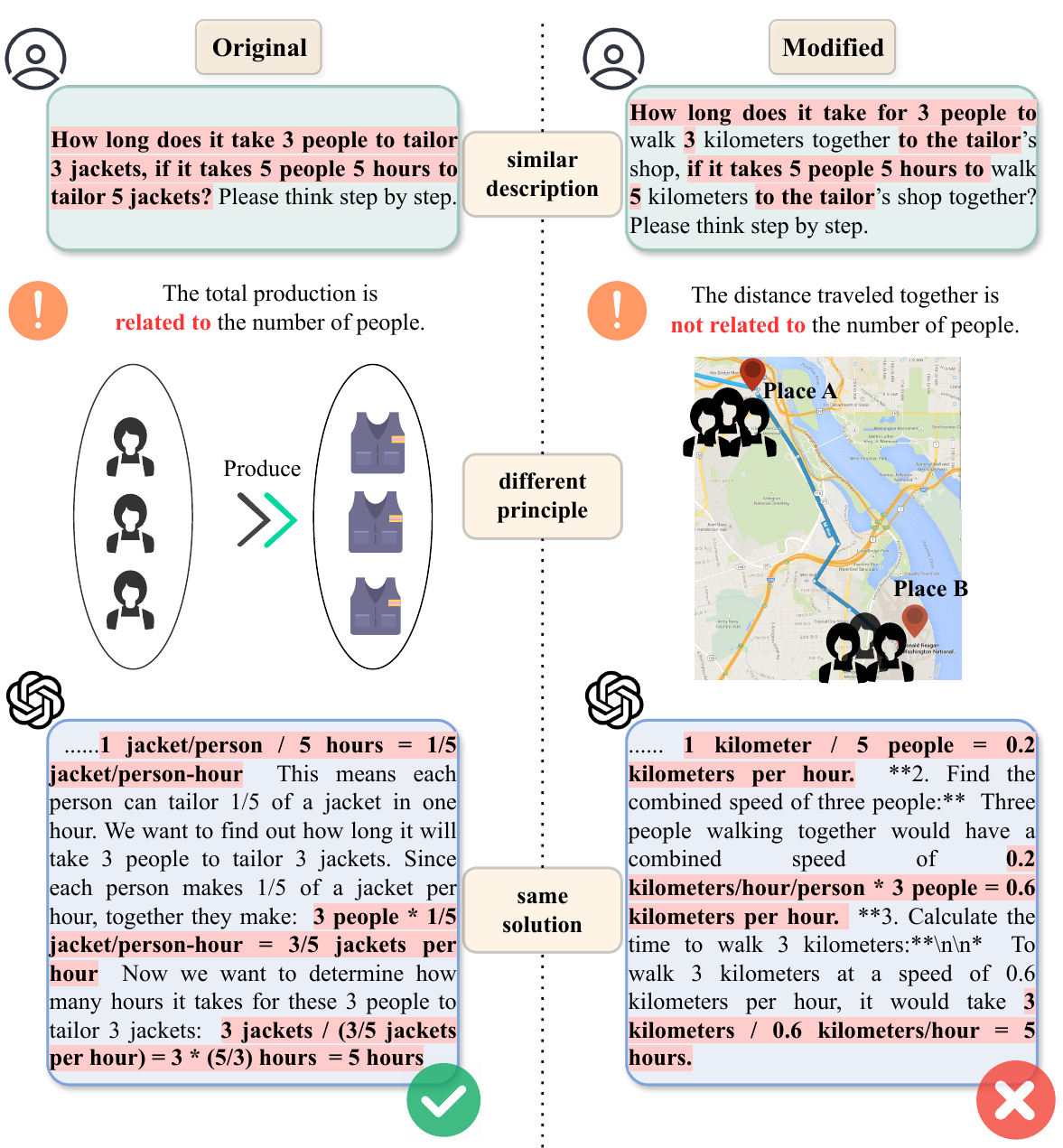}
\caption{An example of an original CRT2 problem and its modified problem.}\label{Fig:fig2-2}
\end{figure}

As illustrated in Fig. \ref{Fig:fig2-2}, we take the first problem in the original CRT2 dataset as an example. In the original problem, all workers work together, and the total amount of product is related to the number of workers. In the modified problem, working together is changed to walking together. Therefore, the total distance walked is independent of the number of people. We calculated the average similarity using the same function as in the previous sub-experiment, yielding 82.33\%. 

\begin{figure}[H]
\centering
\includegraphics[width=0.48\textwidth]{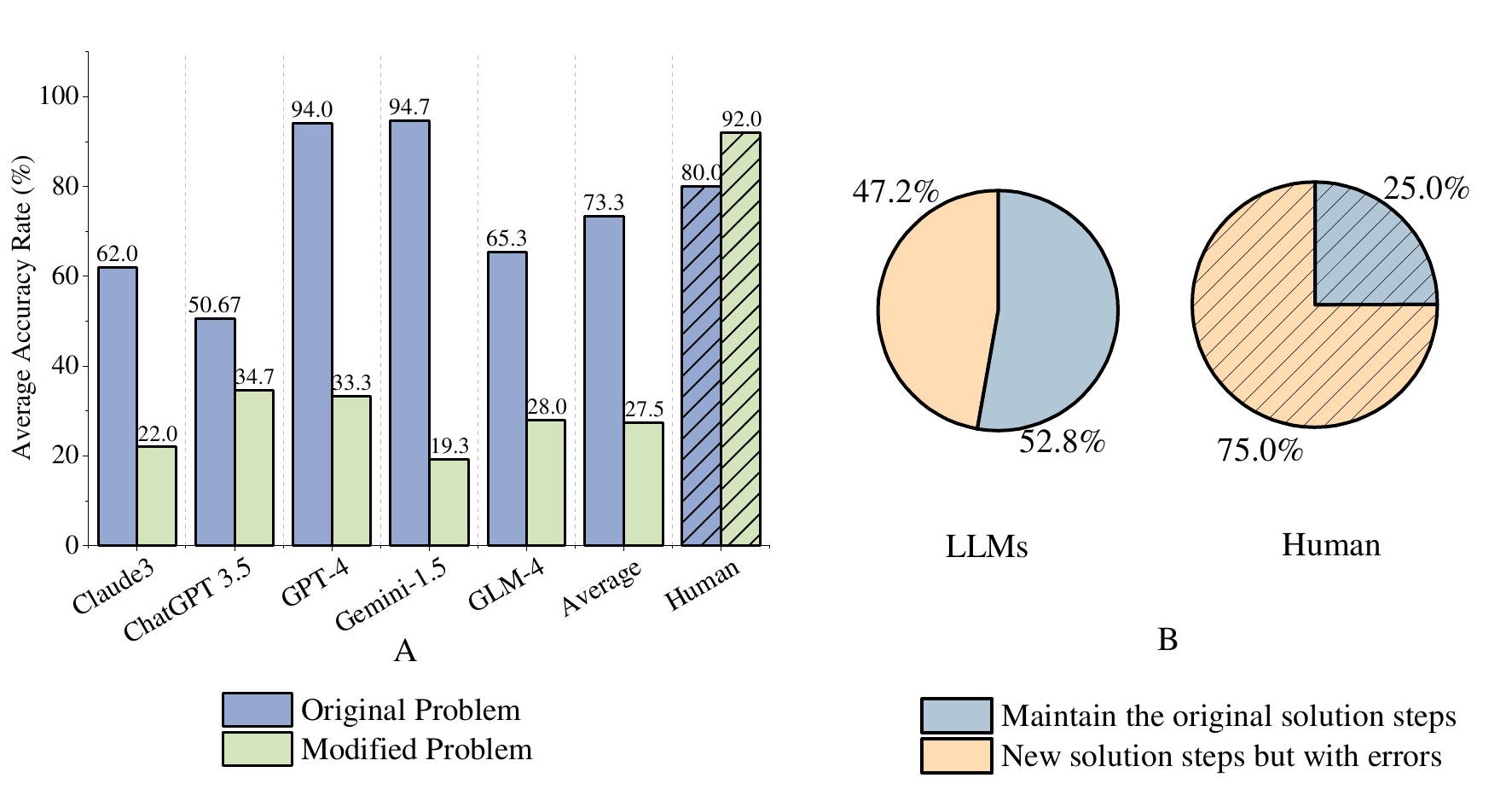}
\caption{Accuracy of the LLMs and Human(with slash) when answering the CRT2 problems (A), and the proportion of incorrect answer types in modified CRT2 problems (B).}\label{Fig:fig_result_2-2}
\end{figure}

As illustrated in Fig.\ref{Fig:fig_result_2-2}A, the experimental findings demonstrate a significant decrease in the accuracy of various models. Claude3’s accuracy plummeted from an initial level of 62.0\% to 22.0\% ($\delta = 40.0\%; \chi^2(1) = 14.82; P < 0.001$). Similarly, GPT-4 experienced a dramatic reduction from 94.0\% to 33.3\%($\delta = 60.7\%; \chi^2(1) = 37.20; P < 0.001$). Gemini-1.5 experienced the most substantial decline, dropping from 94.7\% to 19.3\%($\delta = 75.4\%; \chi^2(1) = 54.85; P < 0.001$). GLM-4 also encountered a considerable decrease, with its accuracy falling from 65.3\% to 28.0\%($\delta = 37.33\%; \chi^2(1) = 12.54; P < 0.001$). ChatGPT 3.5 declining from 50.7\% to 34.7\%($\delta = 16.0\%; \chi^2(1) = 2.00; P = 0.157$). Human experiments showed that the problems became easier.

Upon scrutinizing all erroneous answers to the modified problems, as shown in Fig. \ref{Fig:fig_result_2-2}B, it was observed that 52.8\% of errors employed the original CRT2 methodology to address the modified problems (incorporating the number of participants into the calculation process). This indicates that, due to the similarity in wording between the two types of problems, LLMs sometimes overlook the differences in their mathematical principles and consequently choose the same solution steps as those for the original problems, which are incorrect.

\subsection{For CRT3: Transforming Exponential Growth Problems into Linear Growth Problems}\label{subsec3_3}

\begin{figure}[htp]
\centering
\includegraphics[width=0.48\textwidth]{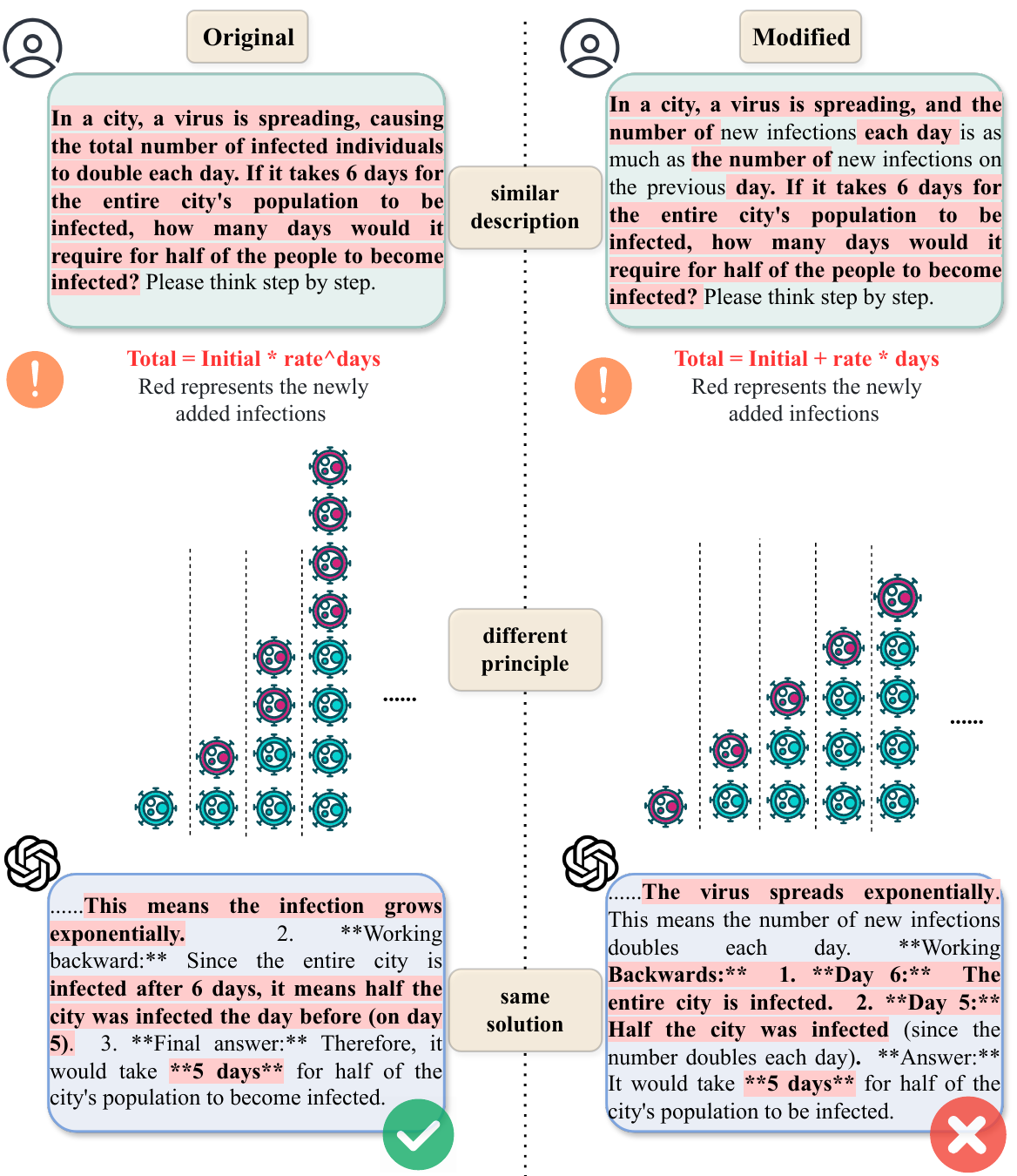}
\caption{An example of an original CRT3 problem and its modified version.}\label{Fig:fig2-3}
\end{figure}

Fig. \ref{Fig:fig2-3} presents the first problem in the CRT3 dataset as an illustrative example. In the original problem, the number of viruses doubles each day, resulting in exponential growth of the total number. In contrast, the modified problem features a constant daily increase in the number of viruses, which leads to linear growth. We calculated the average similarity using the same function, yielding a result of 88.89\%.

\begin{figure}[H]
\centering
\includegraphics[width=0.48\textwidth]{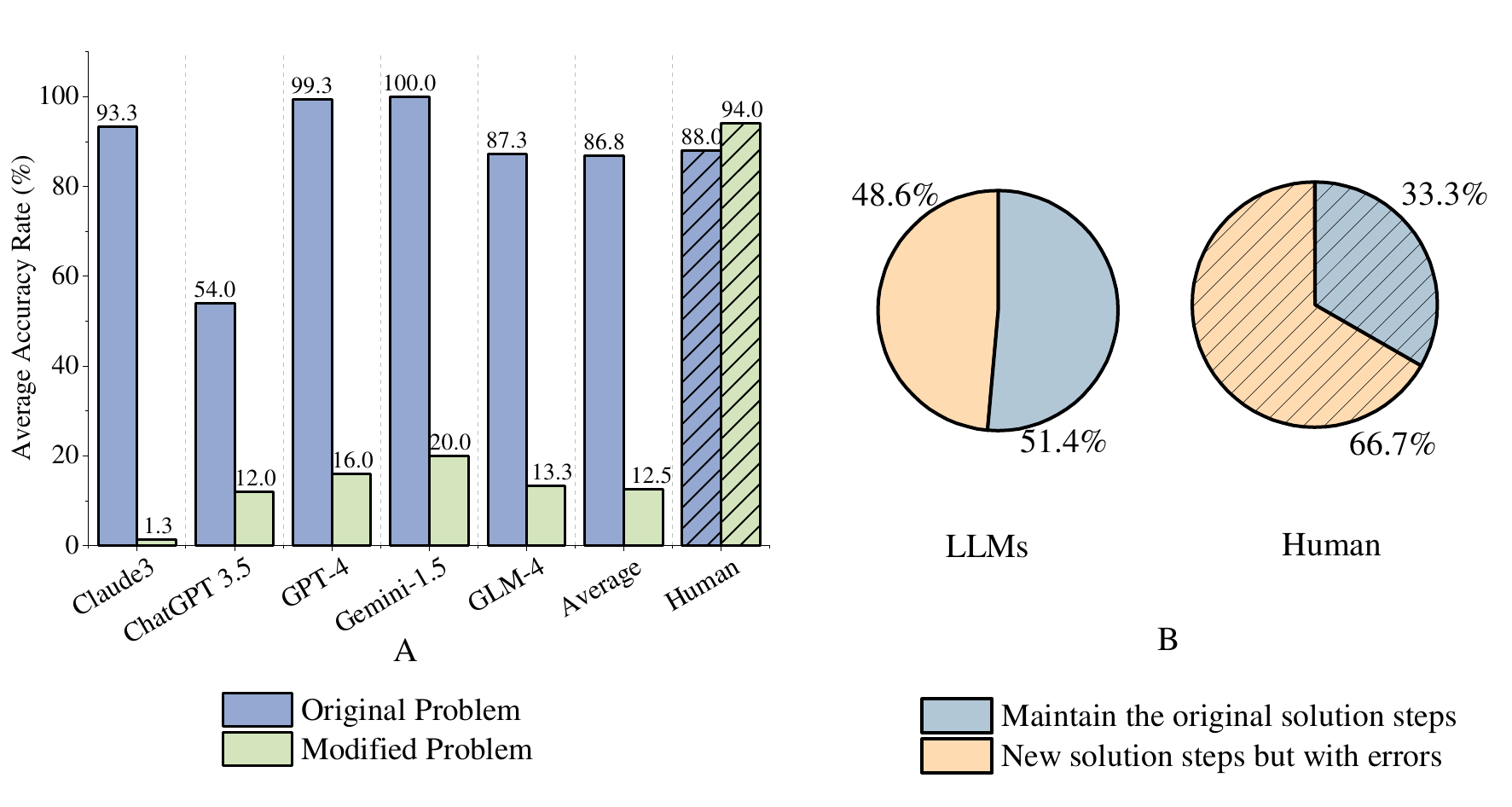}
\caption{Accuracy of the LLMs and Human(with slash) when answering the CRT3 problems (A), and the proportion of incorrect answer types in modified CRT3 problems (B).}
\label{Fig:fig_result_2-3}
\end{figure}

The experimental results in Fig. \ref{Fig:fig_result_2-3}A demonstrate that Gemini-1.5 achieved the highest accuracy of 100\% for the original CRT3 problems. In contrast, ChatGPT 3.5 recorded the lowest accuracy at 54.0\%, with an average accuracy of 86.8\%. Following the modification of the problems, the average accuracy decreased significantly to 12.5\% ($\delta = 64.3\%; \chi^2(1) = 272.83; P < 0.001$). The accuracy rate of Gemini-1.5 decreased to 20.0\%($\delta = 80.0\%; \chi^2(1) = 63.38; P < 0.001$), while Claude3’s accuracy was merely 1.3\% ($\delta = 92.0\%; \chi^2(1) = 81.23; P < 0.001$). Conversely, humans demonstrated enhanced accuracy attributable to the reduced complexity of transforming the initial exponential problems into linear formats.

The main reason for the decline in the average accuracy is that these models continued to employ the problem-solving approach of the original problem when confronted with the modified problem. The proportions of Claude3, ChatGPT 3.5, GPT-4, Gemini-1.5, and GLM-4 that identified the modified problem as one of exponential growth were 81.7\%, 77.8\%, 85.7\%, 72.5\%, and 71.5\%, respectively, although it should have been identified as a linear growth problem. Among these, the proportion of problems that were solved entirely using the original method—yielding the same answer as the original problem—constituted 37.8\%, 31.8\%, 82.7\%, 55.7\%, 52.2\%, and the average is 51.4\% (as depicted in Fig. \ref{Fig:fig_result_2-3}B). This result indicates that more than half (51.4\%) of the errors were due to maintaining the original solution steps.

\textbf{Result Analysis of Experiment II:} The modifications to CRT1, CRT2, and CRT3 altered the mathematical principles underlying these problems to preserve the similarity of their statements. The experimental findings reveal that LLMs sometimes rely on their inherent problem-solving methodologies, even when provided with CoT prompts as guidance. This observation further supports our hypothesis that LLMs can address mathematical problems, likely due to the inclusion of analogous problems in their training datasets. Consequently, LLMs select solutions based on the superficial similarity of problem descriptions rather than demonstrating an understanding of the underlying mathematical principles.

\section{Experiment III: Replicating Experiment I \& II on the Latest o1 Model}\label{sec4}

\begin{figure}[htp]
\centering
\includegraphics[width=0.48\textwidth]{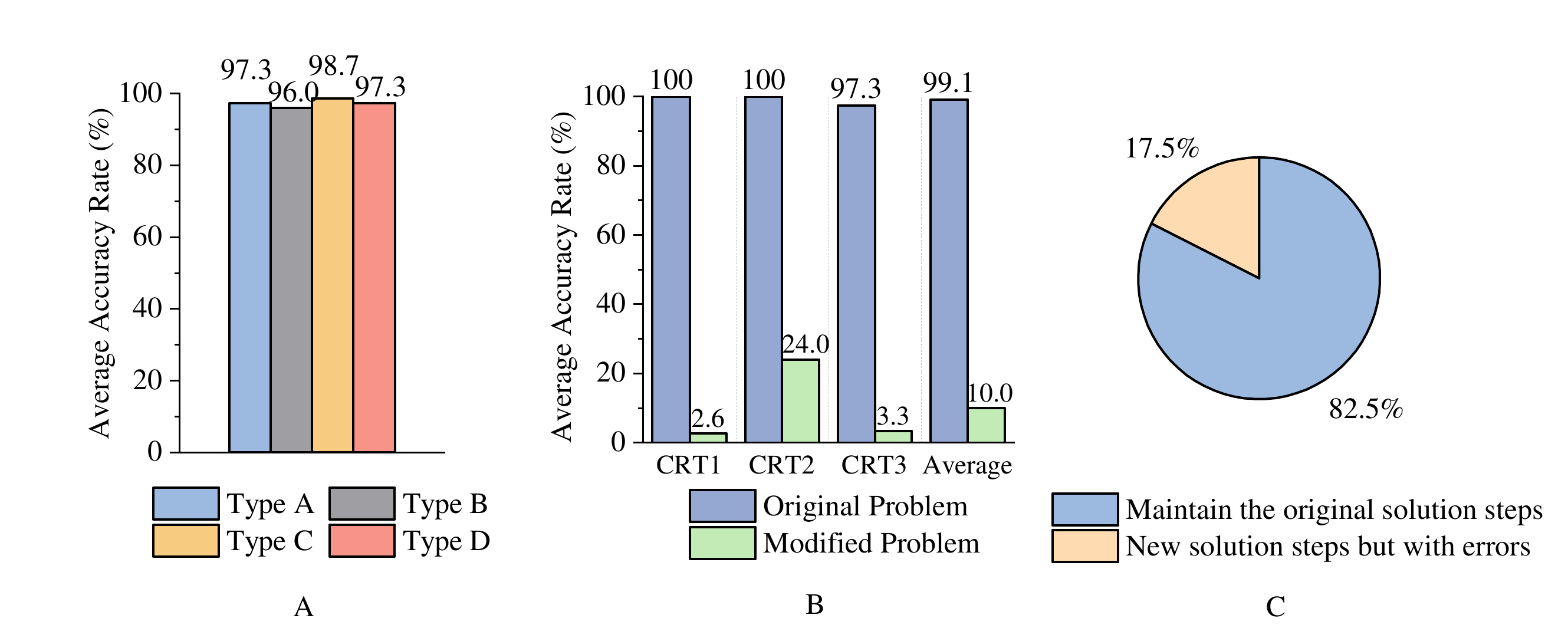}
\caption{Average Accuracy of the o1 model in Experiment I (A) and Experiment II (B), and the average proportion of incorrect answers in Experiment II (C).}
\label{Fig:fig_result_3-1}
\end{figure}

Given OpenAI’s release of the o1-preview model, which focuses on enhancing logical reasoning capabilities(\cite{o1_system_card}), we have repeated the above experiments with it. The results are shown in Fig. \ref{Fig:fig_result_3-1}.

In terms of replicating Experiment I (Fig. \ref{Fig:fig_result_3-1}A), changing the numbers in the problem statements did not significantly affect the accuracy of o1's answers. This observation suggests that the specific number given in the problem statement does not affect the method that o1 selects for solving the problem. This might imply that o1 has potentially incorporated prompts like ``list the equations before solving'' into its built-in thought process. Nevertheless, this speculation cannot be officially confirmed, as OpenAI has not released technical details regarding o1's improved reasoning capabilities.

However, when replicating Experiment II (Fig. \ref{Fig:fig_result_3-1}B), the average accuracy of o1 was only 10.0\%. After the mathematical principles of the CRT problems are altered while modifying the textual description as little as possible, o1 persisted in selecting problem-solving approaches based on the mathematical principles corresponding to the original problems. The errors stemming from this persistence constituted 100\%, 68.4\%, and 75.7\% of the incorrect answers to the three types of CRT problems, respectively. 

As shown in Fig. \ref{Fig:fig_result_3-1}C, on average, 82.5\% of the incorrect answers were due to treating the modified problem as the original one and providing the same answer as for the original problem. This may indicate that technologies such as CoT, prompt engineering, and even fine-tuning tailored to specific datasets (e.g., the o1 model) cannot fundamentally enhance the ability of LLMs to comprehend mathematical problems. The reason for this lies in the fact that the learning paradigm of LLMs has not undergone substantial changes, such as the adoption of auto-regressive next-token predictors, resulting in their thought patterns being deeply ingrained, akin to human intuition (System 1) rather than logical reasoning (System 2).

\section{Related Works}
Some previous empirical research has questioned the reasoning abilities of LLMs(\cite{wu-etal-2024-reasoning,mirzadeh_gsm-symbolic_2024,zhang_careful_2024,sprague_cot_2024}), particularly in solving mathematical problems.  In contrast, the contributions of this paper are more pronounced in the following aspects:

Most of the previous related studies were published before the release of the o1 model (e.g.\cite{wu-etal-2024-reasoning,zhang_careful_2024,sprague_cot_2024}), therefore did not assess this model, which is renowned for its reasoning abilities. The latest experimental evidence provided in this paper demonstrates that similar conclusions remain valid for the newly released o1 model.

Although some previous studies have found that some LLMs exhibit overfitting and memorization effects during the reasoning process, leading to a decline in performance on modified datasets, the degree of decline was generally not significant (e.g.\cite{zhang_careful_2024} , less than 13\%), and not all LLMs experienced a decrease. Consequently, these studies were cautious in drawing conclusions, given that even for human testers, a moderate decline is inevitable. However, in this paper, the accuracy of all LLMs drops by more than 50\% on average, providing more convincing experimental evidence. We utilized System 1 and System 2 theories from cognitive psychology to examine the incorrect responses produced by LLMs. Our analysis revealed a notable decrease in the accuracy of these responses, which contrasts sharply with the performance of human System 2.



Some previous studies have also mentioned human cognitive science to compare and analyze the thinking patterns of LLMs(\cite{wu-etal-2024-reasoning,mirzadeh_gsm-symbolic_2024}.) However, their datasets used for testing mathematical reasoning ability are purely mathematical and unrelated to cognitive science itself. In contrast, our dataset itself is a mathematical test dataset from cognitive science, originally used to study the roles of human reasoning (System 2) and intuition (System 1). Therefore, this paper offers greater interpretability and inspiration from a psychological perspective.

\section{Conclusion}\label{sec13}

This paper draws on the classic CRT problems from human psychology to conduct an empirical study on the ``emergence" of mathematical capabilities in mainstream LLMs. It aims to test whether LLMs possess mathematical reasoning capabilities similar to human System 2 and to provide more interpretable explanations for the causes from the perspective of human psychology. By constructing forward experiments (Experiment I) and reverse experiments (Experiment II), we obtained conclusions that are starkly different from mainstream views. Specifically:


\begin{itemize}
    \item LLMs tend to match problem-solving strategies based on textual similarity rather than truly understanding the underlying principles of mathematical problems. This process is more akin to human intuition (System 1) rather than logical reasoning (System 2).

    \item Even with the introduction of CoT or specialized training to enhance reasoning abilities (e.g., o1), such methods cannot fundamentally alter the problem-solving thinking patterns of LLMs to endow them with System 2-like rational logical reasoning abilities. The emergence of this phenomenon is perhaps attributable to the dominant paradigm utilized in the training and fine-tuning of LLMs, which fundamentally involves predicting the next token with the highest probability. This mechanism closely aligns with human intuition (System 1), which enables rapid decision-making based on patterns correlated with high probability, thereby presenting a certain distinction from System 2.
\end{itemize}

This study conducts an empirical analysis of the ``emergence" of LLMs' mathematical reasoning abilities from a psychological perspective. We hope that this study can reduce overblown expectations of LLMs' capabilities and stimulate more empirical research to objectively evaluate the current limitations of LLMs' abilities. 




\balance

\printbibliography

\end{document}